\documentclass[10pt,a4wide]{article}

\usepackage[pagebackref,breaklinks,colorlinks]{hyperref}
\usepackage{rotating}

\usepackage{graphicx}
\usepackage{amsmath}
\usepackage{amssymb}
\usepackage{amsfonts}
\usepackage{booktabs}
\usepackage{comment}
\usepackage{multirow}
\usepackage{array}
\usepackage{xcolor}
\usepackage{bm}
\usepackage{pifont}

\usepackage{tikz}
\usepackage{algorithmic}

\newcommand{\softsign}{\mathrm{softsign}}

\title{How to make effective use of domain experts for image classification?}

\author{Dieu-Donné Fangnon$^1$, Diane Lingrand$^1$, Aurélie Liard$^2$, \\
  Marco Corneli$^{1,2}$, Antoine Pasqualini$^2$, Frédéric Precioso$^1$\\
  $^1$ Université Côte d'Azur, Inria, Maasai Team, Sophia Antipolis, France\\
  $^2$ CEPAM, CNRS, Université C\^ote d'Azur, France\\
  {\tt\small firstname.lastname@univ-cotedazur.fr}}

\begin{document}
\maketitle
\begin{abstract}
A lot of expectations have been put for years on integrating domain expert knowledge in image classification models. Several approaches have been explored, Concept Bottleneck Models (CBMs) opened up a new avenue of research leading to many variants, and more recently to Concept-based Embedding Models (CEMs). CBM consider binary encoding of each concept, while CEM expands this idea by embedding each concept through two vectors. However in real-life scenarii, domain experts' knowledge is usually organized in concepts determined by various attributes, each attribute encoded either with numerical values, or range of values, or binary values, or categorical values. In this work, we first finetune an image feature extractor for classifying attributes representing the downstream object classes, where the class attributes have been specified by experts under various encoding formats. A classification head is then learnt from these various attributes to categorize target objects. We experimentally show that it improves the classification for three datasets: Kaggle fish dataset, AWA2 and a more challenging new wood charcoal dataset. We then propose an automatic selection of potential missclassified data. In this second step, experts are asked for those data to eventually modify the predicted attributes in order to improve the classification.
\end{abstract}

\section{Introduction}
\label{sec:intro}

\begin{comment}
\color{red}[A FAIRE: réécrire (ou réorganiser) l'introduction (jusqu'à 'Contributions') en ne commençant pas par les charbons mais plutôt:
\begin{itemize}
    \item Dans ce travail on souhaite utiliser les connaissances des expert.e.s des domaines visés en 2 interventions:
    \begin{itemize}
        \item en amont lors de la définition des attributs
        \item en aval avec la correction de certains attributs prédits
    \end{itemize}
    \item Se positionner par rapport aux approches basées sur les concepts (qui ne considèrent que les critères binaires)
    \item Dire qu'on a étudié un dataset classique des approches concepts (AWA2) mais qu'on a aussi travaillé sur un dataset plus complexe en terme d'attributs (les poissons) et que l'application concrète ce sont les Hard Wood charcoals.
\end{itemize}\color{black}
\end{comment}

%%% FRED %%% \mc{Neuro-symbolic AI \citep{bhuyan2024neuro} is an active research field, whose aim is to develop hybrid AI models that better exploit the human expertise by directly injecting it into the learning procedure. This work falls within that context and, more precisely, it relies on the two following cornerstones: concept-based models ~\citep[CBM,][]{pmlr-v119-koh20a} and concept embedding models~\citep[CEM,][]{NEURIPS2022_867c0682}.}

Finding an effective way for injecting human expertise into the learning procedure while preserving impressive performances of deep learning models is a challenge. Concept Bottleneck Models (CBMs)~\cite{Chauhan23} classify objects in images through a human-readable intermediate layer based on predefined attributes (or concepts) of these objects. It is straightforward to interpret since this intermediate layer encodes each concept with a binary value (i.e. presence, yes/no), but this choice hinders reaching high accuracy. To solve this limitation, in Concept Embedding Models (CEM), each concept is represented by two vectors~\cite{NEURIPS2022_867c0682} (one encoding the presence of the concept and one encoding its absence), at the cost of clear interpretation. Concepts have since been used to foster both interpretability and accuracy~\cite{Gupta2024ASO}. Although explanation is not always guaranteed~\cite{Mahinpei2021PromisesAP}, CEMs provide also the user with interacting and correcting the model~\cite{schramowski20,Chauhan23,bontempelli2023conceptlevel}. 

One limitation of the approaches cited so far is in the nature of the expert criteria (or attributes), their embedding is homogeneous (either all binary, or all embedded into two vectors) while in real-life scenarii human experts provide attributes in very diverse format. Moreover, the relevance of attributes is class-dependent: inside some classes, all data share the exact same values of a part of the attributes' set (present or absent) while the remaining attributes are not relevant for this class because they can be present or absent. One of the main contributions of this work is to introduce more general loss functions allowing one to work with heterogeneous expert attributes and varying number of attributes per class. Moreover, we exploit the domain experts' knowledge in an original way through two interventions: \emph{a priori}, in the definition of the class attributes and \emph{a posteriori} with the correction of some predicted attributed.

We evaluate our approach on three datasets of images provided with domain experts criteria: the well known AWA2 dataset~\cite{xian2018zero} as well as on the more challenging Kaggle fish dataset\footnote{\url{https://www.kaggle.com/datasets/markdaniellampa/fish-dataset/data}}~\cite{fish_dataset_2022}, including heterogeneous attributes (categorical data, integers, intervals) related to morphological properties of the fishes. Then, we focus on the taxonomic identification of Hard Wood charcoals from microscopic image, using a new real dataset provided by archaeologists which has not being cleaned up or prepared for AI methods or concept-bases approaches as AWA2 did. Taxonomic identification of charcoal is of particular interest for the bio-archaeological community and a dedicated section of the paper (see Sec.\ref{sec:hwdataset}) better defines the associated issues.
Here, we just point out that, while observing at microscope the three anatomical sections of charcoal fragments (transversal, tangential and radial), typical criteria used by wood anatomists for the identification are the presence/absence of scalariform perforation plates, uniseriate rays, vessels in radial multiples, etc. That argument is conducted in a standardised vocabulary, the IAWA list of microscopic attributes~\cite{wheeler1989iawa}, which underpins the InsideWood database~\cite{wheeler2011insidewood} and is the operational language of wood Indexation worldwide. As for the Kaggle fish dataset, the IAWA attributes are heterogeneous.

\smallskip
In summary, this paper brings the following main contributions.
\begin{enumerate}\itemsep2pt
    \item We extend the concepts or class attributes types from classic CBM dataset (AWA2~\cite{xian2018zero}
    ) by adding new types encountered in real applications and defined by domain experts: binary values that could also be variable, set of sizes, numerical values or intervals and categories and propose adapted loss functions (Sections \ref{sec:attributes} and \ref{sec:encodingAndLosses}).
    \item We improve the classification results by using these class attributes(Section \ref{sec:exp}).
    \item We propose an algorithm of automatic selection of potential misclassified samples and a scheme of interaction with experts to correct some of the attributes in order to improve the classification (Section \ref{sec:postProc}).
  %  We formulate charcoal identification as multi-task learning over a standardised expert anatomical vocabulary, with a faithful differentiable encoding of its three annotation regimes: a masked criterion loss that abstains on expert-declared variability, and a normalised soft interval-membership penalty for ordinal magnitudes, both in closed form (\cref{sec:exp}).
  %  \item We contribute a leakage-free hardwood charcoal benchmark of $16$ species, $393$ specimens and $3{,}313$ micrographs, in which five genera contain multiple congeneric species, with per-species IAWA annotations scored by an expert on carbonised material (\cref{sec:dataset}).
  %  \item We report the first criterion-level evaluation of a charcoal identification model, how often the predicted anatomy agrees with the expert's with a qualitative audit performed by a domain specialist.
  %  \item Unlike the CUB-200 settings, our criteria are not purely Boolean but also categorical or interval-valued, on far fewer images of a different nature. Predicting them first, then classifying either from the fine-tuned descriptor or from the predicted criteria, improves accuracy in both cases over the attribute-free baseline, on all three datasets (\cref{sec:other}).
\end{enumerate}

\section{Three image based classification problems and associated attributes}
\label{sec:attributes}
\begin{figure}[!htb]
  \includegraphics[width=\linewidth]{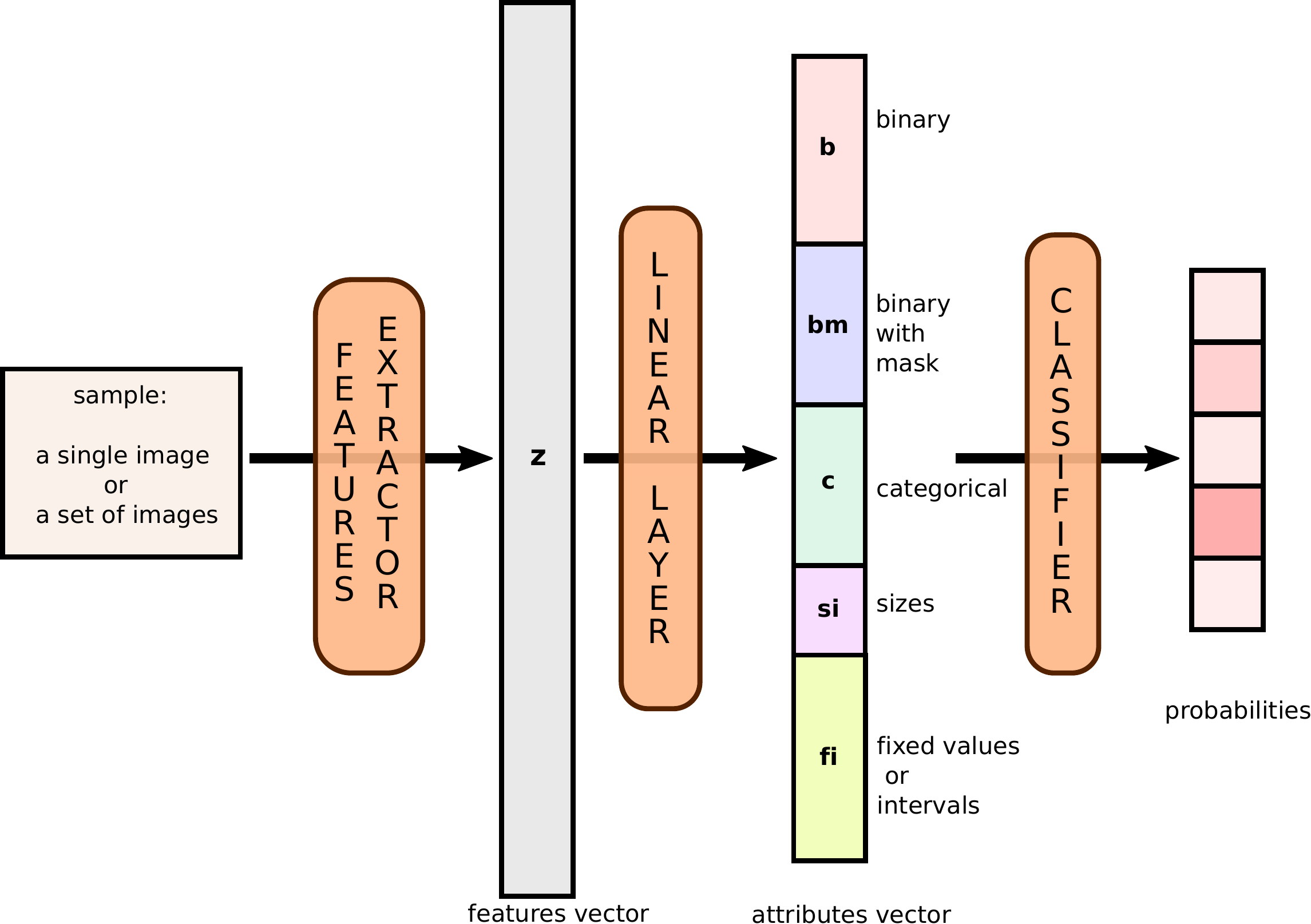}
  \caption{The global picture of this paper. The data are single images or set of images. We extract features for the attributes prediction. Then, classes are predicted from attributes.}
  \label{fig:globalSch}
\end{figure}

Figure~\ref{fig:globalSch} presents the main architecture of the paper that will be applied for our 3 classification problem. In the case of AWA2 and the Kaggle fish dataset, each data is represented by a single image while in the hard wood charcoal dataset, each data (or specimen) is represented by a set of images. From data, a feature extractor computes the feature vector from which the attributes are predicted using a linear layer with dedicated activations. The feature extractor can be (part of) a CNN, a visual transformer or any other model. The last layers of this feature extractor are then finetuned for predicting the experts' attributes for each class. Then, classes are predicted from the attributes vector. 

Although a direct classification from feature vectors with attributes fine-tuning would lead to a similar accuracy, our scheme offers opportunities for later interactions with domain experts.

\subsection{AWA2 dataset}
In many works on concepts, each data or class can be described using concepts or attributes that are binary as it is the case in the \textbf{AWA2 dataset}. Each AWA2 class is described using 85 binary attributes concerning the physical appearance of animals but also their habitat, behavior and feading. 

\subsection{The Kaggle fish dataset}
The Kaggle fish dataset is based on the fish dataset\footnote{\url{https://www.kaggle.com/datasets/markdaniellampa/fish-dataset/data}}~\cite{fish_dataset_2022} using attributes from the fishbase search engine\footnote{\url{fishbase.se}}~\cite{fishbase}. Even with the information that this dataset is about fishes from the Marinig Fishing Port in Cabuyao City, The Philippines, some species names could not be identified clearly in fishbase or were not present. We thus limited our study to 12 species (a total of 5715 data), and collected anatomic properties from fishbase. These attributes are listed on table \ref{tab:speciesAndAttributes}: some are fixed values, some are interval of values and some other categories.

\begin{center}\begin{table}
    \centering{
\begin{tabular}{|c||c|c|c|c||c|c|}
    \hline
    Name & \begin{sideways}Dorsal spines\end{sideways}& \begin{sideways}Dorsal soft rays\end{sideways} & \begin{sideways}Anal spines\end{sideways} & \begin{sideways}Anal soft rays \end{sideways} & \begin{sideways}Body shape\end{sideways} & \begin{sideways}Cross section\end{sideways}\\
      \hline
    \hline
    Bangus & 2 & 13-17 & 2 & 8-10 & e & ov\\
    \hline
    Big Head & 3 & 7 & 1-3 & 12-14 & f & ov\\
    Carp &&&&&& \\
    \hline
    Fourfinger & 9 & 13-15 & 3 & 14-16 & e & co\\
     Threadfin &&&&&& \\
    \hline
    Gourami & 16-20 & 7-10 & 9-11 & 8-11 & f & co\\
     &&&&&  & \\
    \hline
    Grass & 0 & 10-11 & 0 & 10-14 & c & ci\\
    Carp &&&&&& \\
    \hline
    Indo-Pacific & 0 & 16-21 & 0 & 23-31 & e & co\\
    Tarpon &&&&&& \\
    \hline
    Knifefish & 9-10 & 12-13 & 2-3 & 11-12 & e & co \\
        \hline
    Mosquito & 0 & 7-9 & 0 & 9-10 & f & ci \\
    Fish  &&&&& &\\
    \hline
    Mudfish & 0 & 38-43 & 0 & 23-27 & e & ot \\
    \hline
    Silver & 4 & 8 & 3 & 6-7 & f & ov\\
    Barb &&&&& &\\
    \hline
    Silver & 0 & 7-10 & 0 & 11-17 & f & ov \\
    Carp  &&&&& &\\
    \hline
    Tenpounder & 0 & 20-27 & 0 & 14-18 & e & ov\\
    \hline
\end{tabular}}
\caption{Table of attributes available for each fish specie. Four first attributes are values or range of values. Body shapes are categorical variables: elongated (e), fusiform/normal (f) or cylindrical (c), mapped to [0,1,2]. Cross sections are also categorical: oval(ov), compressed (co), circular (ci) or other (ot), mapped to [0,1,2,3].}
\label{tab:speciesAndAttributes}
\end{table}
\end{center}
\subsection{The hard wood charcoal dataset} \label{sec:hwdataset}

This dataset is composed by 393 specimens from 16 hardwood species (grouped into 3 families and 10 genera). For each tree, we have a set of images from radial, tangential and transverse sections, with a variable number of images. These images are high-resolution micrographs of burned wood (charcoal). The purpose of the classification is to classify specimens and not single images.

An international set of visual attributes have been established for fresh wood identification by the IAWA organisation~\cite{InsideWood,wheeler2011insidewood}. Anthracologists (i.e. archaeologists who study ancient wood charcoal) have adapted this list of attributes to our case of \emph{burnt} wood and selected 86 anatomical attributes they consider as informative, still denoted by IAWA attributes in the paper. The complete list of IAWA attributes we consider in this paper is available in the appendix (table~\ref{tab:intervals}).
The IAWA attributes, per class, can be \emph{always present}, \emph{always absent} or \emph{variable}, i.e. they may be present or absent.  
As we will see in the following, in order for the loss function not to penalize or emphasize those values, they are masked.
Moreover, some of the IAWA attributes are concerning sizes and we grouped them together in order to predict a single real value that is however constrained in a given range: i)
\textbf{intervessel pit size} (crit. 24 to 27): Minute ($\leq 4 \mu m$), Small (4-7), Medium (7-10) or Large ($\geq 10 \mu m$); ii)
\textbf{mean tangential diameter of vessel lumina} (crit. 40 to 43): below 50~$\mu$m, 50-100, 100-200 or above 200~$\mu$m; iii)
\textbf{vessels per square millimeter} (crit. 46 to 50): below 5, 5-20, 20-40, 40-100 or above 100.

The final IAWA attributes representation is thus composed of 73 binary (potentially variable) attributes and 3 size attributes.

\paragraph{Context and related works.}

Since charcoal preserves the cellular structure of the wood it was made from, it carries a direct record of the vegetation that past societies gathered and burned~\cite{theryparisot2010anthracology}. For this reason, its taxonomic identification is of interest for archaeologists.  

Automatic identification from wood micrographs has progressed from handcrafted descriptors~\cite{martins2012database} to convolutional networks~\cite{ravindran2018classification,wu2021wood}, with the consistent finding that combining the three sections outperforms any single plane~\cite{rosa2022improved,zielinski2025advanced}. Charcoal, instead, has received far less attention~\cite{maruyama2018automatic,oliveira2021automatic} and its identification is strictly harder: carbonisation shrinks, cracks and vitrifies tissues, and shifts quantitative measurements enough that size classes established on fresh wood no longer transfer directly~\cite{zemke2025variation}. Reviews of the field~\cite{hwang2021review,ravindran2022caveat} identify two open problems, protocol rigour and interpretability. We take the first as settled and adopt unchanged the stratified specimen-level protocol and the permutation-invariant set encoder~\cite{zaheer2017deepsets} of~\cite{fangnon:hal-05660387}, so that any change in performance is attributable to the supervision we add.

On \emph{fresh wood}, detectors trained on annotated micrographs already localise IAWA features and support identification an anatomist can follow~\cite{liu2025interpretable}. These results establish that the vocabulary is learnable, but their conditions do not hold for charcoal: 
carbonisation degrades the structures themselves~\cite{zemke2025variation}, so criteria must be scored on carbonised material rather than inherited from fresh-wood references, and expert annotation is available only per species. To our knowledge, IAWA criteria have not been used as supervision for a charcoal identification model, and the agreement between predicted and expert-annotated criteria has not been reported for carbonised material.
\begin{comment}
\smallskip

We therefore place the expert vocabulary inside the model, attaching to the fused specimen representation heads that reproduce, in differentiable form, what the annotation actually asserts, which is where the technical content lies, because that annotation is not a clean binary matrix. It marks some criteria as \emph{variable} within a species, so we \emph{abstain} on those through a masked objective rather than inject label noise on exactly the features that separate close relatives. Three groups of criteria are ordered magnitude bins rather than independent attributes, so we predict one scalar per group under a penalty flat inside the expert's admissible interval and rising outside it, supervising the band the expert asserts rather than a point estimate they never provided.
\end{comment}

\section{Encoding the attributes and loss functions}
\label{sec:encodingAndLosses}
In this section, we describe the expert attributes vectors as well as their different components (binary, binary with mask, size, values, intervals, categories as depicted in figure \ref{fig:globalSch}). Moreover we specify how their prediction is integrated in the model via activation and loss functions.

\subsection{Binary values (b)}
% AWA2
In AWA2, all attributes are binary values (present / absent) that can be predicted using a sigmoïd activation and binary cross-entropy loss (BCE) or mean square error (MSE). 

\subsection{Binary values with mask (bm)}
% AIWOOD
When a binary attribute is marked as \emph{variable} for one class, we introduce a mask in order to ignore it (for that class).
We had to deal with such a scenario in the hard wood charcoal dataset, where not all attributes were clearly present or absent, depending on the class. For example, attribute \#12 (Solitary vessel outline angular) has to be present for class \textit{Erica cinerea}, absent for class \textit{Sorbus aria} but is variable for class \textit{Rosa canina} (i.e. it can be present or absent). Since the number of variable attributes differs from one class to another, we divide the loss component concerning all non variable attributes by the number of non variable attributes instead of the classic total number of attributes. This keeps the loss on a comparable scale across different species.
Without doing that, species with many variable criteria would contribute systematically smaller gradients.

\subsection{Sizes seen as intervals (si)}
\label{sec:interval}
\begin{figure}
    \centering
    \includegraphics[width=0.9\linewidth]{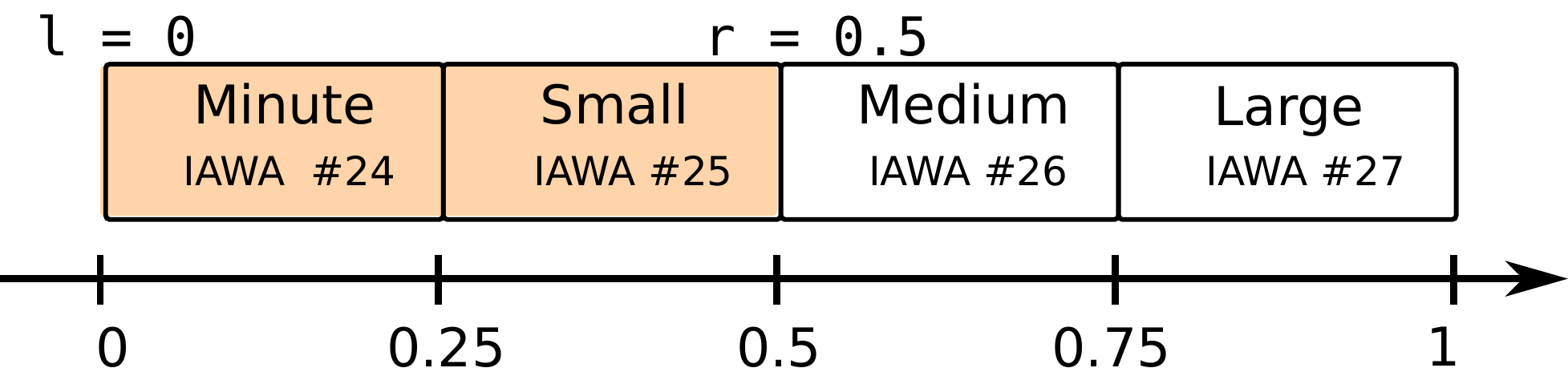}
    \caption{Example of the IAWA feature ``intervessel pit size'' with 4 different ordered values as described in section~\ref{sec:hwdataset}. The unit interval is divided in 4 equal parts. Left (l) and right (r) values from equation~\ref{eq:iawafn} depend on the \emph{present} value(s), represented here by coloured boxes: IAWA \#25 and \#26 are always present for the class considered, so that left value is 0 and right value is 0.5.}
    \label{fig:oneIAWAinterval}
\end{figure}

Among the AIWA criteria, some are about intervals, as for example the 
attribute ``intervessel pit size'' (section \ref{sec:hwdataset}). As it can be seen in figure~\ref{fig:oneIAWAinterval}, it is made of several sub-intervals, each one possibly being labeled as \emph{always present}, \emph{absent} or \emph{variable} for different classes.
The presence/absence of each sub-interval in not exclusive among species. For the species considered in figure~\ref{fig:oneIAWAinterval}, for example, the leftmost intervals are both present (coloured boxes).

The approach we adopted is to map all possible values to the unit interval $[0,1]$, that we split uniformly according to the number of possible sizes (4 in the example we are considering). 
For each class, we convert the \emph{present} sub-intervals (always contiguous) to an admissible interval $I_c^b=[\,l_c^b,\,r_c^b\,]$ for species $c$ and IAWA attribute $b$. Once more, the illustration of this mapping is presented on figure~\ref{fig:oneIAWAinterval}.

\begin{figure}
  \includegraphics[width=\linewidth]{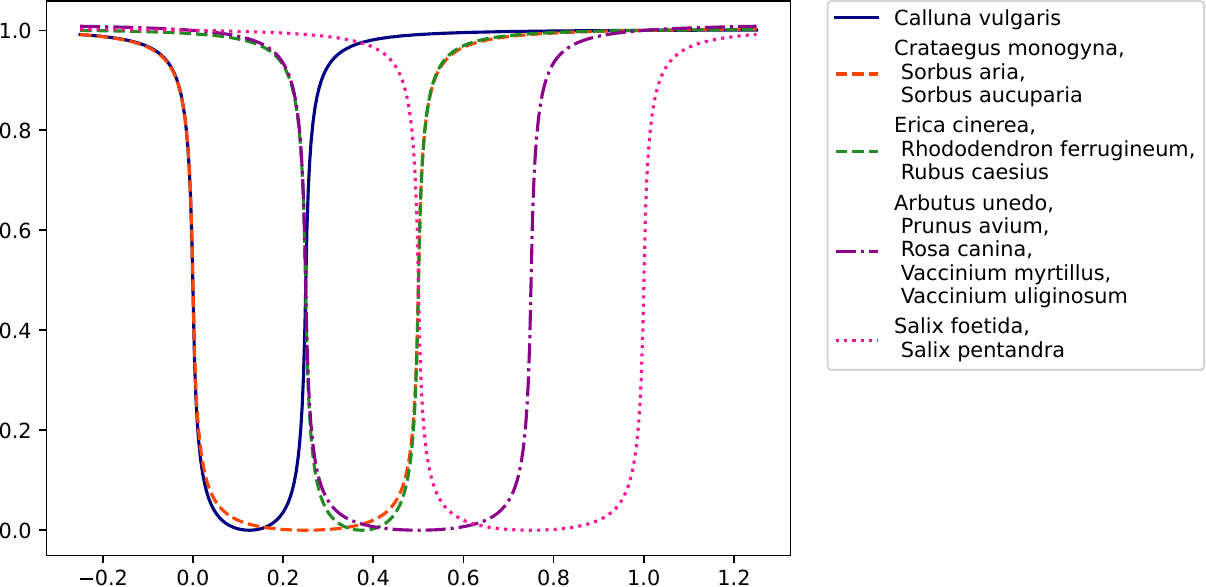}
  \caption{The different interval values functions concerning the 'intervessel pit size' for the different species. In this case, many classes share the same function for this anatomical attribute because they share the same intervals. The goal of this loss function is to be minimal in the correct interval.}
  \label{fig:plotFnsIps}
\end{figure}

Then, we need a penalty that vanishes inside $I_c^b$, saturates outside it, and is differentiable everywhere. Writing $\sigma_\lambda(t)=\softsign(\lambda t))$ for a smooth step of sharpness $\lambda$ ($\lambda=100$, an hyperparameter in this approach), the function:
\begin{equation}
f_{si}(c,b,u)=1-\sigma_\lambda\!\bigl(u-l_c^b\bigr)+\sigma_\lambda\!\bigl(u-r_c^b\bigr)
\label{eq:iawafn}
\end{equation}
forces the predicted values to be inside an admissible interval $I_c^b$ (see figure~\ref{fig:plotFnsIps} for the functions concerning the intervals from IAWA\#24 to IAWA\#27). We rescale it to $f_c^{\,b}:[0,1]\!\to\![0,1]$ so that the penalty is comparable across bands of different width and position. The loss is effectively forcing the result of the previous function to be close to zero:
\begin{equation}
\mathcal{L}_{\text{si}}=\sum_{b=1}^{3} \frac{1}{C} \sum_{c}\left( \frac{1}{|c|}\sum_{\,i:\,y_i=c}
\bigl[-\log\bigl(1-f_{si}(c,b,u_{ib})\bigr)\bigr]\right)
\label{eq:lint}
\end{equation}
where $C$ is the total number of class and $|c|$ the number of data in class $c$, $u_{ib}$ is the prediction of the $b$ attribute for data $i$ belonging to class $c$ and the inner expectation is over specimens of class $c$ in the batch and the outer over classes present in it, making equation~\ref{eq:lint} a macro-average that balances an imbalanced dataset.

The design choice worth emphasising is that $\mathcal{L}_{\text{si}}$ is \emph{flat} inside $I_c^b$. The expert asserts a range, never a point value: regression towards a bin centre would supervise information the annotation does not contain, and independent per-bin classification would allow $\le\!50\,\mu\mathrm{m}$ and $\ge\!200\,\mu\mathrm{m}$ to fire at once. Equation~\ref{eq:lint} is a smooth surrogate for the likelihood of an interval-censored observation, and asserts exactly what the expert asserts.

\subsection{Fixed values or intervals (fi)}
\begin{figure}
  \includegraphics[width=\linewidth]{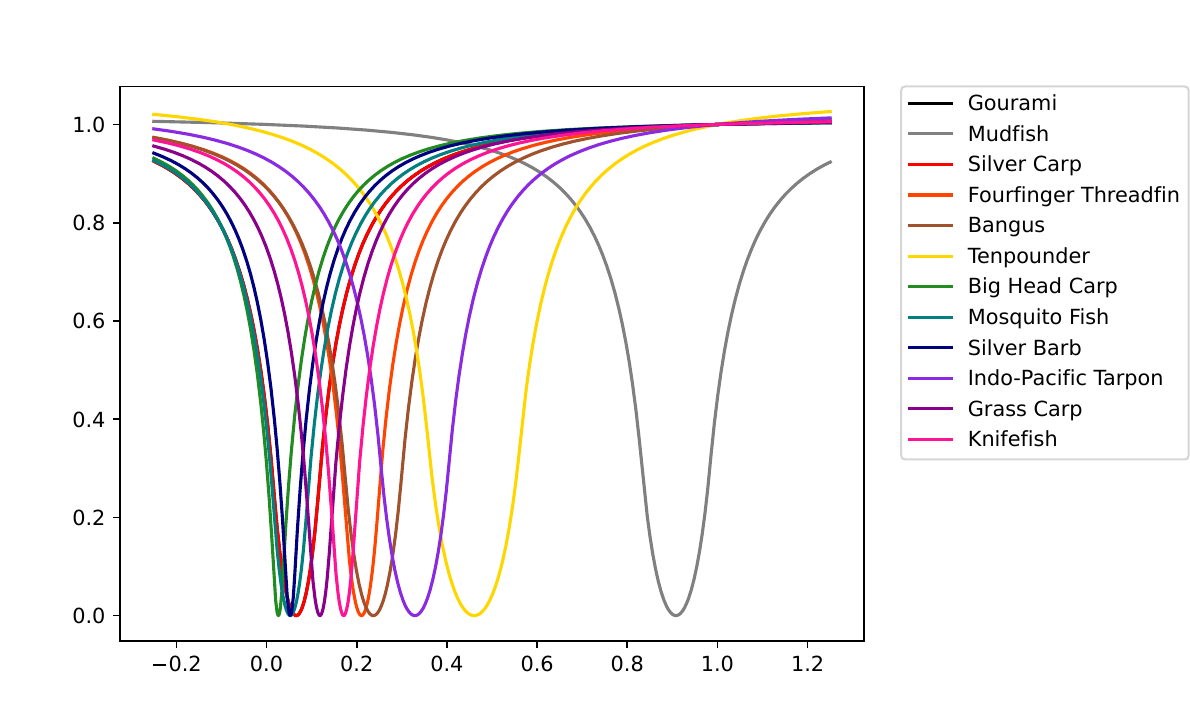}
  \caption{The interval values functions ``Dorsal soft rays'' for different species. Values for ``Dorsal soft rays'' have been rescaled between 0 and 1. The goal of this loss function is to be minimal in the correct interval.}
  \label{fig:plotFnsDsr}
\end{figure}

% les intervals plus libres dans Fish
In table~\ref{tab:speciesAndAttributes}, the four first attributes (Dorsal/Anal spines/soft rays corresponding to $b\in \{0,1,2,3\}$) can be fixed to either a single value or to a range of values. We have converted each value $u$ to a simple interval $[u\  u]$. Similarly to section~\ref{sec:interval}, we built a set of functions composed by a combination of 2 \textit{softsign} functions for allowing an interval of values (or a single one) and is defined as follows for each of the $b$ attributes Dorsal/Soft spines/rays:
\begin{eqnarray}
f_{fi}(c,b,u)= 1 -  \sigma_\lambda\!\bigl(u-l_c^b-\epsilon\bigr)+\sigma_\lambda\!\bigl(u-r_c^b+\epsilon\bigr)
\end{eqnarray}
where $\epsilon=\frac{1}{100}$ is necessary when left $l_c^b$ and right $r_c^b$ interval borders values are the same (single value). Left ($l_c^b$) and right ($r_c^b$)  values are the normalised values from table~\ref{tab:speciesAndAttributes}. An illustration of those functions for ``Dorsal soft rays'' is presented in figure~\ref{fig:plotFnsDsr}. What differ in those functions along the attributes Dorsal/Anal spines/soft rays are the interval of rescaling. The hyperparameter $\lambda=10$ has been chosen for the fish dataset experiments.
The corresponding loss is similarly defined as:
\begin{equation}
\mathcal{L}_{\text{fi}}=\sum_{b=1}^{3} \frac{1}{C} \sum_{c}\left( \frac{1}{|c|}\sum_{\,i:\,y_i=c}
\bigl[-\log\bigl(1-f_{fi}(c,b,u_{ib})\bigr)\bigr]\right)
 \end{equation} in order to force the result of the anatomical function to be as small as possible so that the $u$ values belong to their interval defined on table~\ref{tab:speciesAndAttributes}.

\subsection{Categories (c)}
% dans Fish
Some attributes can be categorical. For example, in the fish dataset, ``Body shape'' and ``Cross section'' are categorical values and will be encoded accordingly (see figure~\ref{fig:anatPropFishes}). Class cross-entropy is used for each of categorical values.

\subsection{Combining all losses}
\label{sec:objective}

\begin{figure}
  \includegraphics[width=\linewidth]{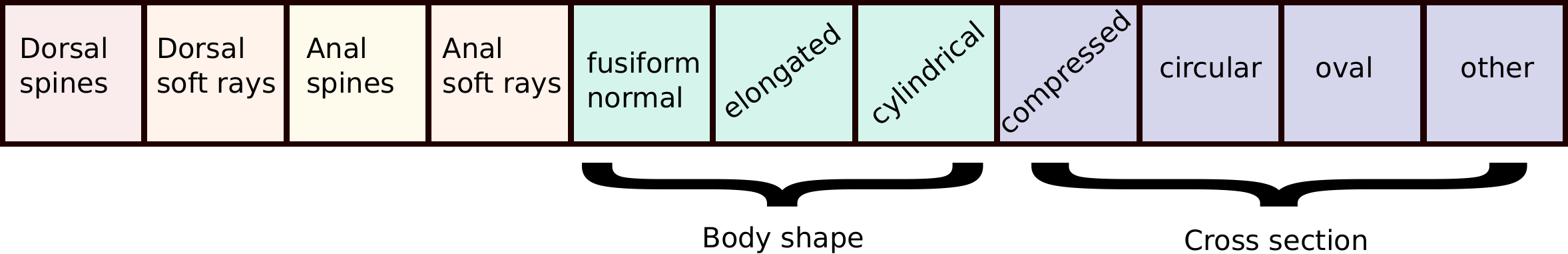}
  \caption{The vector of anatomy of fishes: 4 float values (rescaled between 0 and 1) and 2 sets of probabilities for ``Body shape'' and ``Cross section'' types.}
  \label{fig:anatPropFishes}
\end{figure}

When different types of attributes are present, we need to weight the different loss functions described so far. That is the case for the fish dataset (see figure~\ref{fig:anatPropFishes}) and the hard wood charcoal dataset (the description vector is composed of 73 binary attributes and 3 sizes attributes).

The different objectives differ in scale, noise level and in how much of the specimen representation they can plausibly constrain, so fixed weights would require an expensive grid search. We instead learn a homoscedastic uncertainty per loss~\cite{kendall2018multitask}, with learnable log-variances $\bm{s}=(s_b, s_{bm}, s_{si}, s_{fi}, s_c)$ initialised at zero:
\begin{equation} 
\mathcal{L}_{\text{combined}} =\sum_{j\in\{b, bm, si, fi, c\}}
\Bigl(e^{-s_j}\,\mathcal{L}_j + s_j\Bigr),
\label{eq:iawa}
\end{equation}
where all losses are not mandatory but will be selected according to the classification task to be chosen.
The first term down-weights losses the model finds intrinsically noisy; the second prevents the degenerate solution $s_j\!\to\!\infty$. The $s_j$ are optimised jointly with the network parameters. 

\section{Improving classification using attributes prediction fine-tuning}
\label{sec:exp}

\subsection{Fish classification}
\subsubsection{Baseline}
The classification baseline has been computed using an EfficientNetV2, large size, pre-trained on ImageNet1k as a feature extractor. The classification layer was changed according to the new number of classes. The backbone was frozen except the last 20 layers. The convergence was obtained after 18 epochs based on an early stopping criterion on the CE validation loss (patience 10, delta P$10^{-3}$), using AdamW optimizer (default PyTorch parameters). The test accuracy of the baseline is of 98.96\%

\subsubsection{Prediction of anatomical properties}
The prediction vector is composed of the 4 values of ``Dorsal spines'', ``Dorsal soft rays'', ``Anal spines'' and ``Anal soft rays'' followed by the probabilities of classes of ``Body shape'' and ``Cross section'' (figure~\ref{fig:anatPropFishes}). 
As a first step, we verified that the 4 predicted values were belonging to the correct intervals and that the Body Shape ad Cross-section categories were correct (see figure~\ref{fig:classPredictedValues}).

\subsubsection{Classification from anatomical properties}
\begin{figure}
  \includegraphics[width=0.9\linewidth]{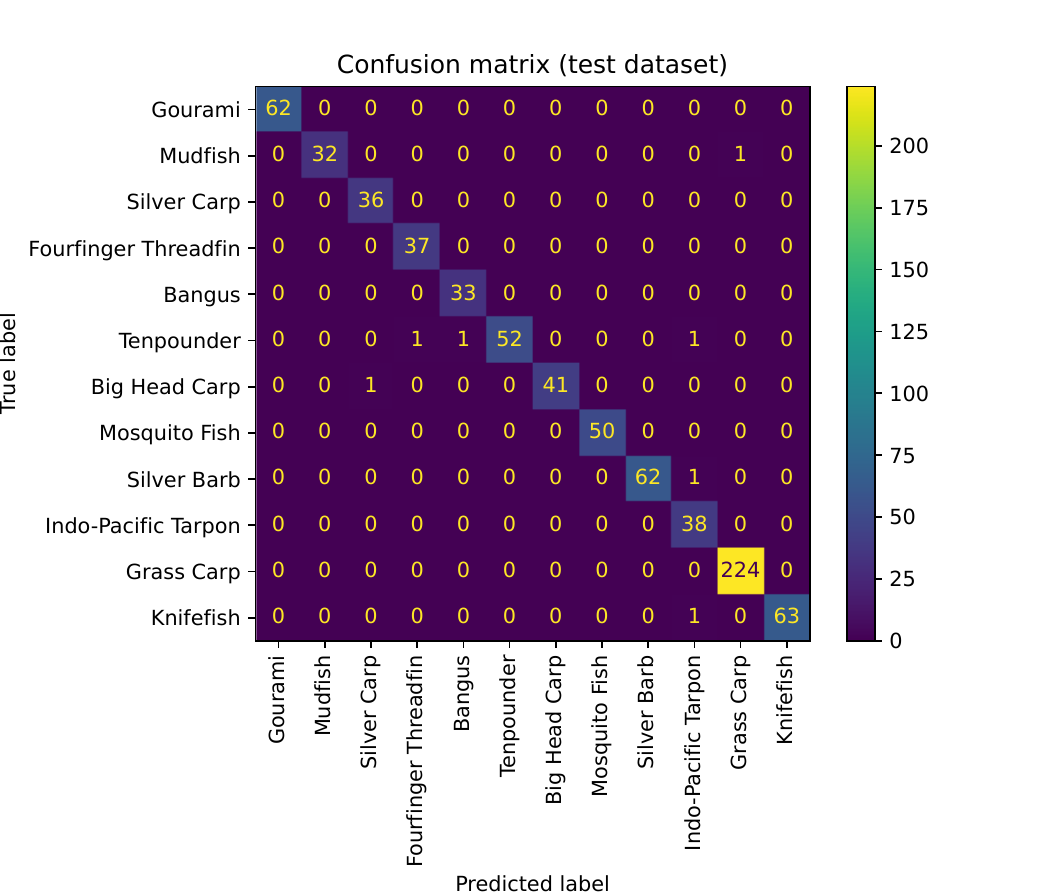}
  \caption{Confusion matrix for the classification of fish species from their anatomical features predictions.}
  \label{fig:cm-predFromAnat}
\end{figure}

The final classifier is composed by a single linear layer (input dimension 11, output dimension 12). Learning was done using classical AdamW optimizer, early stopping on validation loss (patience 10, delta  $10^{-3}$). The test accuracy has increased to 99.22 \% (see confusion matrix on figure~\ref{fig:cm-predFromAnat}).

\subsection{Animals with Attributes 2 (AWA2)}

We considered a subset of 13 classes from the \textit{Animal with Attributes 2 (Awa2)} dataset: Grizzly bear, Killer whale, Dalmatian, Horse, German shepherd, Skunk, Tiger, Moose, Spider monkey, Fox, Hamster, Rhinoceros and Rabbit (a total of 10703 images). Each class is associated with 85 binary attributes: some are always observable (fur, color\ldots) and some rarely because they relate to food, habitat or behavior.

We performed a similar experiment: the baseline is composed by an EfficientNetV2, large size, pre-trained on ImageNet1k and a classification layer for the 13 classes. The prediction layer and the last 20 features layers are learned using AdamW optimizer, early stopping on validation loss (patience 10, delta  $10^{-3}$) leading to a test accuracy of 95.6\%. The prediction of attributes is done similarly with a prediction layer of 85 neurons mean square loss. Then, the attributes are forwarded to a linear layer for classification (13 classes). Test accuracy of the final classification is 97.07\% which is slightly better than the baseline.

\subsection{Hardwood charcoal classification}
\label{sec:setup}

This dataset shares properties with the one analysed in~\cite{fangnon:hal-05660387}: as said in section~\ref{sec:hwdataset}, each tree is represented by a variable number of images taken along three anatomical sections and a quite small number of specimens populates each class. This leads us to represent each specimen by the same architecture based on EfficientNet-V2L~\cite{tan2021efficientnetv2} and a fusion network that finally outputs a single feature vector per specimen. The EfficientNet-V2L backbone is frozen while the fusion part is fine-tuned. 

The input specimens are partitioned by stratified group $5$-fold cross-validation. All results are reported in terms of mean and standard deviation specimen-level accuracy across the five outer folds.

\subsubsection{Baseline results and improvement using IAWA features predictions} Similarly to the experiments with the Kaggle fish dataset and AWA2, we compare the baseline (i.e feature extractor followed by a linear layer for class predictions and classification) performance consisting of  $80.94\%$ of accuracy (at the species level) with the classification from the  predicted IAWA features (as depicted in figure~\ref{fig:globalSch}), resulting into a $82.05\%$ of accuracy.

\subsubsection{Classifier improvement}
\label{par:classifierWood}
The original classifier is a linear layer but can be replaced by any other classifier. We experimented 3 other classifiers: $k$-nearest-neighbour classifier ($k=8$, euclidean metric), multinomial logistic regression (LogReg), and a linear SVM ($C=1$). Results are reported in table~\ref{tab:main2}.
The classification is improved using the logistic regression to 83.19\% of accuracy.

\begin{table}[t]
\centering
\small
\setlength{\tabcolsep}{4pt}
\begin{tabular}{@{}lcccc@{}}
\toprule
% Method & linear layer & $k$-NN & Logistic regression & linear SVM \\
Method & linear layer & $k$-NN & LogReg & SVM \\
\midrule
Baseline  & 80.94$\pm$0.05 & 73.80$\pm$0.02 & 82.69$\pm$0.00 & 81.94$\pm$0.01 \\
With IAWA & 82.05$\pm$0.01 & 75.83$\pm$0.02 & \textbf{83.19$\pm$0.04} & 79.39$\pm$0.06 \\
\bottomrule
\end{tabular}
\caption{\textbf{Specimen-level accuracy (\%) on the 16-species benchmark},
mean $\pm$ std over five stratified folds, with EfficientNet-V2L features at
$500\!\times\!500$, best per feature block in bold. Comparison of different classifiers.} 
\label{tab:main2}
\end{table}

\begin{figure}[t]
    \centering
    \includegraphics[width=\linewidth]{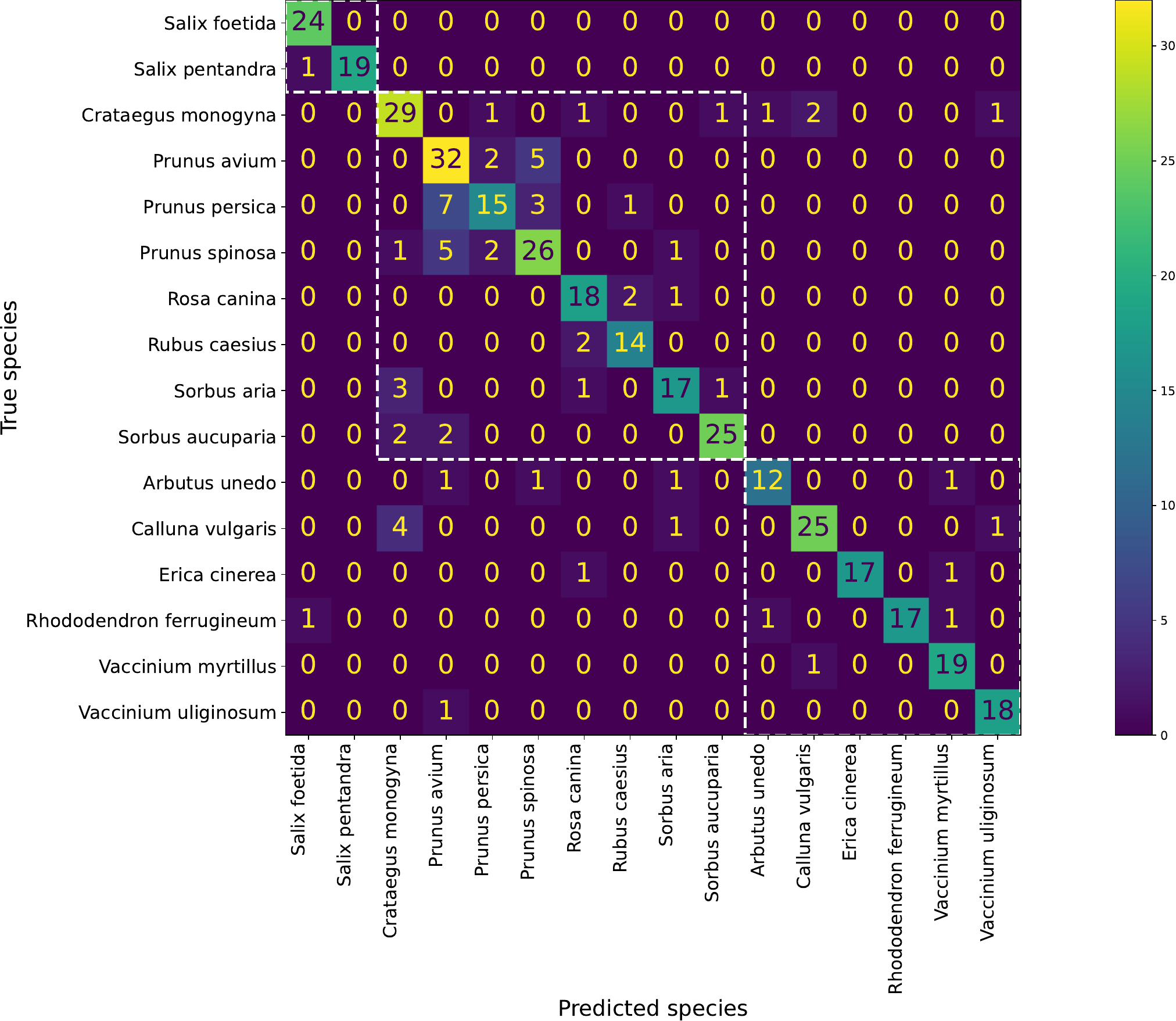}
    \caption{\textbf{Species-level confusion matrix} for the best configuration, IAWA feature, logistic regression; $83.19\pm0.04\%$), pooled over the five test folds. White boxes mark the families \emph{Salicaceae}, \emph{Rosaceae} and \emph{Ericaceae} from top-left to bottom-right. The accuracy at family level accuracy is $\mathbf{96.18\%}$.}
\label{fig:cm-iawa}
\end{figure}
\subsubsection{Classification at the \emph{species } or \emph{family} level}
\label{par:species-families}
Wood experts classify trees into species, inside genus and inside families. In our case, 3 families are present: Salicaceae, Rosaceae and Ericaceae. 
In figure~\ref{fig:cm-iawa}, we observe that errors remain taxonomically local: family-level accuracy is $96.18\%$. Of the $66$ misclassifications, $26$ are congeneric (i.e. inside the same genus), $24$ among the three \emph{Prunus} species alone, where \emph{P. persica} is the weakest class at $15/26$ with seven specimens assigned to \emph{P. avium}, and a further $25$ cross genus boundaries while staying inside a family, leaving only $15$ cross-family confusions. The residual difficulty is therefore congeneric, as expected of species sharing most of their gross anatomy, and is what the section~\ref{sec:iawacorrecting} addresses. 

\section{Post-processing interaction with experts}
\label{sec:postProc}
\subsection{Interacting with the experts after the learning (case of AWA2 dataset)}
\label{sec:interact}
In an interactive learning process, at inference on new unlabeled samples, we would like to automatically identify the misclassified samples and ask experts to annotate them. Furthermore, we would aim as little and simple interventions on those samples to improve the model. Here we conduct a preliminary study on what could be optimally expected on those two aspects: relevant sample selection and little interventions.\\
\noindent\textbf{\textit{Sample selection criterion}}
For a new unlabeled sample, we compute its attribute vector $u$, then for each target class $c$ we compute $\mathcal{L}_{\text{combined}}(u,c)$ (see Eq. 5) and search for the minimum value. The selection criterion for this sample is thus the minimum of $\mathcal{L}_{\text{combined}}$ losses along classes. 

\begin{figure}
  \centering
  \includegraphics[width=0.8\linewidth]{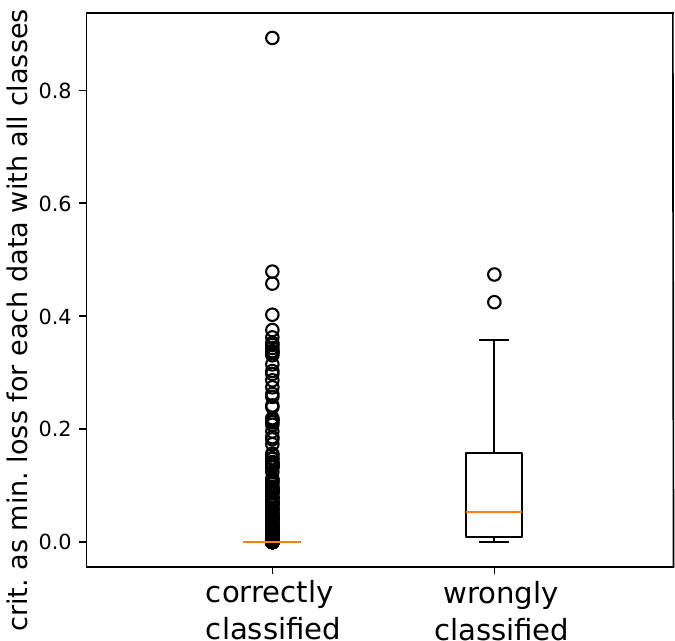}
  \caption{Misclassified samples are usually associated with a higher criterion value than correctly classified once, although a few exceptions exist.}
  \label{fig:histoAWA2MSElossCorrectVsWrong}
\end{figure}

\begin{figure}
  \centering
  \includegraphics[width=0.9\linewidth]{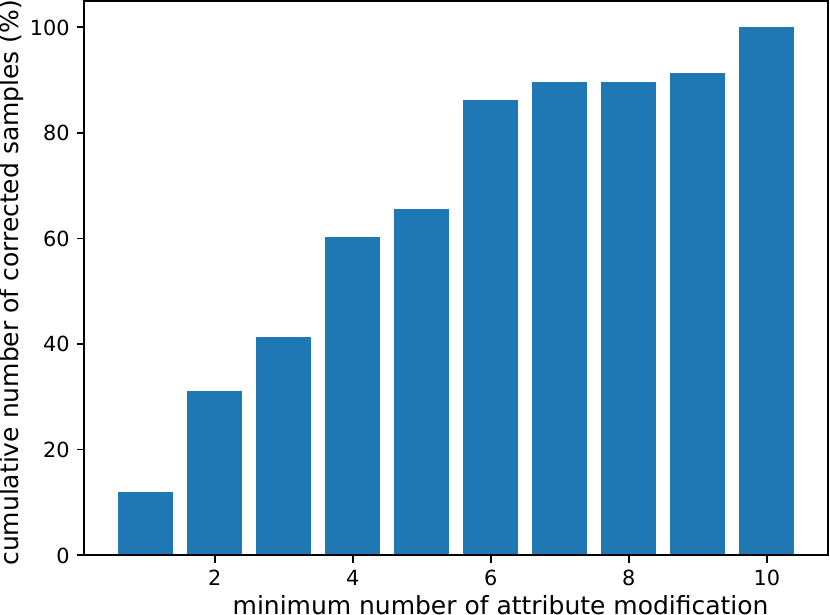}
  \caption{Awa2 dataset: minimum number of attributes to be modified for correct classification of the test set.}
  \label{fig:awa-cumNbAttChg}
\end{figure}

Figure~\ref{fig:histoAWA2MSElossCorrectVsWrong} shows that this criterion is mainly smaller for correctly classified data than for misclassified. In figure~\ref{fig:histoAWA2MSElossCorrectVsWrong}, even if more samples are correctly classified, one cannot see the thickness of the box (IQR: interquartile range) because all correctly classified criterion values are concentrated below 0.01 even if few correctly classified sample outliers can reach higher values than the median of wrongly classified criterion values. Wrongly classified samples present mostly higher values: higher median but also, larger IQR. Thus, we propose to ask experts to examine samples corresponding to the highest criterion values and to check how many of their attributes could be obviously incorrect.
\\
\noindent\textbf{\textit{Amount of interventions}}
More in details, in AWA2, 58 over 1944 samples are misclassified. The sample \#259 presents the highest criterion value and is a dalmatian misclassified into grizzly bear. However, it is possible to correct the classification by modifying only 4 attributes over 85.
Sample \#708 has the smallest criterion (7e-10) over the missclassified (german shepherd classified as spider monkey) but is still the 1347th over 1944 increasing order criterion. The sample \#708 needed the modification of 6 attributes to be classified correctly. However, sample \#1838 corresponds to the 6th bigest loss (0.1128) and is correctly classified. More generally, with no more than 6 attribute changes, the test accuracy can reach 99.6\% and every sample can be corrected using no more than 10 attributes (see figure~\ref{fig:awa-cumNbAttChg}).

It is usually difficult to intervene on AWA2 attributes because they are not all observable: even if few pictures depict a bear eating a salmon, the attribute \textit{''what animals are eating''} is usually not observable in the images. An annotator could also easily assign the attribute \textit{''habitat''} to a rabbit by knowing that a rabbit lives in burrows but not by observing this from the image. 

In opposite, in the hardwood charcoal dataset, the IAWA features are all supposed to be observable. 

\subsection{Anthracologist experts correcting IAWA attributes}
\label{sec:iawacorrecting}

Because anthracologist experts have set the IAWA features we use in this paper, it is natural for them to compare our predictions with their annotations. The analysis in this section is performed using the predictions of the best-performing IAWA model reported in table~\ref{tab:main2}, namely logistic regression, which achieved a mean accuracy of 83.19\% across the five folds.

\begin{figure}[t]
\centering
\includegraphics[width=0.8\columnwidth]{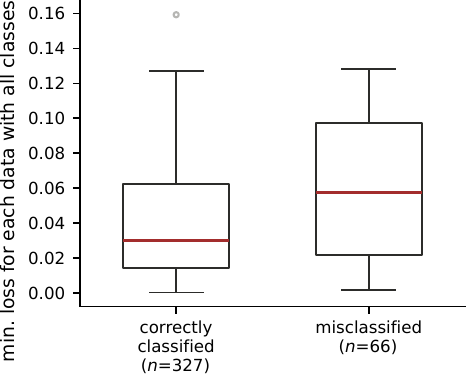}
\caption{Minimal loss between each data and all classes. Misclassified specimens score higher, so the quantity can be used to select which specimens to submit to an expert.}
\label{fig:triage}
\end{figure}

Figure~\ref{fig:triage} shows criterion values (with the same protocol as in section~\ref{sec:interact}) for IAWA attributes with the same conclusion: samples with higher values are the ones to be presented to experts.
Misclassified specimens score roughly twice as high as correctly classified ones (median $0.057$ against $0.030$, figure~\ref{fig:triage}), although the separation is only partial. 
The $25$ worst-scoring specimens of $393$ contain $14$ of the $66$ errors. Thus, a specialist reviewing the flagged portion detects a disproportionate share of the errors.

\begin{figure}[t]
\centering
\includegraphics[width=0.8\columnwidth]{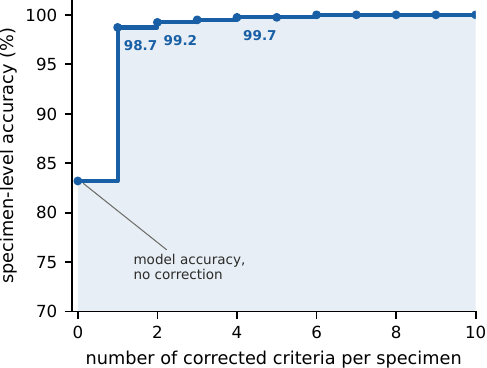}
\caption{Specimen-level accuracy under progressive expert correction of the predicted criteria.}
\label{fig:cum-changes}
\end{figure}

The classification can be improved with only few IAWA attribute corrections. Starting from an accuracy of $83.19\%$, a single correction criterion raises it to $98.7\%$, with $61$ of the $66$ errors solved through six corrections for every specimen in the test set (figure~\ref{fig:cum-changes}). Indeed, a failed identification involves one or two attributes rather than a wholesale misreading of the anatomy, which is what makes expert correction practical: the specialist confirms a single attribute rather than re-examining the fragment. We achieved the same performances on AWA2, where six attribute changes correct $99.6\%$ of samples. 

\paragraph{Expert-in-the-loop intervention.}

To verify that the automated procedure accurately replicated what a specialist would have done for the remaining samples, an anthracologist reexamined eight misclassified samples belonging to different classes and noted, for each attribute on which the model had erred, the value observed for that fragment. Applying between four and eight readings per fragment corrects seven of the eight (table~\ref{tab:expert}), including three congeneric confusions within \emph{Prunus}, the hardest cases in the dataset and one specimen assigned to a different family that four corrections resolve. The single failure is instructive: one correction was supplied for a \emph{Sorbus aucuparia} specimen. This is a first audit on 8 specimens and we report it as such.

Also, even if the IAWA features are all supposed to be observable, there are still cases where some IAWA features are not observable: because the technician did not captured them on images or because some very few are not observable on small branches (only trunc).

\begin{table}[t]
\centering
\small
\setlength{\tabcolsep}{4pt}
\begin{tabular}{@{}lccll@{}}
\toprule
Specimen & \#wrong & \#read & \multicolumn{2}{c}{Identified as, before $\to$ after} \\
\midrule
\texttt{\scriptsize AIERICALCRE}  & 14 & 7 & \emph{Erica} $\to$ \emph{Calluna} & \checkmark \\
\texttt{\scriptsize GEODE}        & 16 & 4 & \emph{Rhododendron} $\to$ \emph{Calluna} & \checkmark \\
\texttt{\scriptsize AIROSPRUEUR75} & 4 & 2 & \emph{P.~avium} $\to$ \emph{P.~spinosa} & \checkmark \\
\texttt{\scriptsize AIROSPRUCREf}  & 8 & 3 & \emph{P.~avium} $\to$ \emph{P.~persica} & \checkmark \\
\texttt{\scriptsize AIERICALEX62} & 17 & 6 & \emph{Crataegus} $\to$ \emph{Calluna} & \checkmark \\
\texttt{\scriptsize AIERICALLAGf} & 12 & 6 & \emph{V.~myrtillus} $\to$ \emph{Calluna} & \checkmark \\
\texttt{\scriptsize AIW24}         & 9 & 8 & \emph{Crataegus} $\to$ \emph{P.~avium}  & \checkmark \\
\texttt{\scriptsize AIROSSORBELa}  & 7 & 1 & \emph{Crataegus} $\to$ \emph{Crataegus} & \ding{55} \\
\bottomrule
\end{tabular}
\caption{\textbf{Expert intervention on eight misclassified specimens.} The first four are \emph{Calluna vulgaris}, then \emph{Prunus spinosa}, \emph{P. persica}, \emph{P. avium} and \emph{Sorbus aucuparia}. \#wrong: criteria the model read incorrectly among evaluable entries; \#read: values recorded by the anthracologist, a subset of those flagged. Identification is by nearest species prototype under the mean squared error of section~\ref{sec:iawacorrecting}. }
\label{tab:expert}
\end{table}

\section{Conclusion and perspectives}
\label{sec:ccl}

Instead of learning a model for class prediction, we learn a model for class attribute prediction that can be interpreted by experts. These attributes are first determined by experts of the application domain and can be of different natures (binary, category, values, intervals\ldots). We experimented with 3 different datasets that predicting attributes inside the workflow of classification improves not only the classical classification metrics but also enables specific interactions with experts. Losses computed for learning the attributes vectors can also be used as a criterion for the selection of data to be reviewed by experts. By simulation of experts modifications, we have shown how to correct all misclassifications by a minimum number of attribute modifications. 
For the last data, hardwood charcoal, we had the opportunity to ask real anthracologists about few misclassified samples and be able to correct them using few attribute changes.

Future work will concern an exhaustive study of those corrections with experts.

\subsubsection*{Data Availability} AWA2 is a public dataset with images and class attributes. Kaggle fish data set is a public images dataset and class attribute are given in table~\ref{tab:speciesAndAttributes}. The modern wood charcoal images dataset supporting the results reported in this paper is available upon request for research purposes. 

\subsubsection*{Aknowledgments}
This work was carried out as part of the AIWOOD project, funded under grant ANR-23-CE38-0013. The authors are grateful to the OPAL infrastructure from University C\^ote d'Azur for providing resources and support.

{
    \small
    \bibliographystyle{plain}
    \bibliography{main}
}

\clearpage

\section*{Supplementary material}
\section{Kaggle fish dataset: range of predicted values}
\label{sm:fishAtt}

\begin{figure}[h]
  \includegraphics[width=\linewidth]{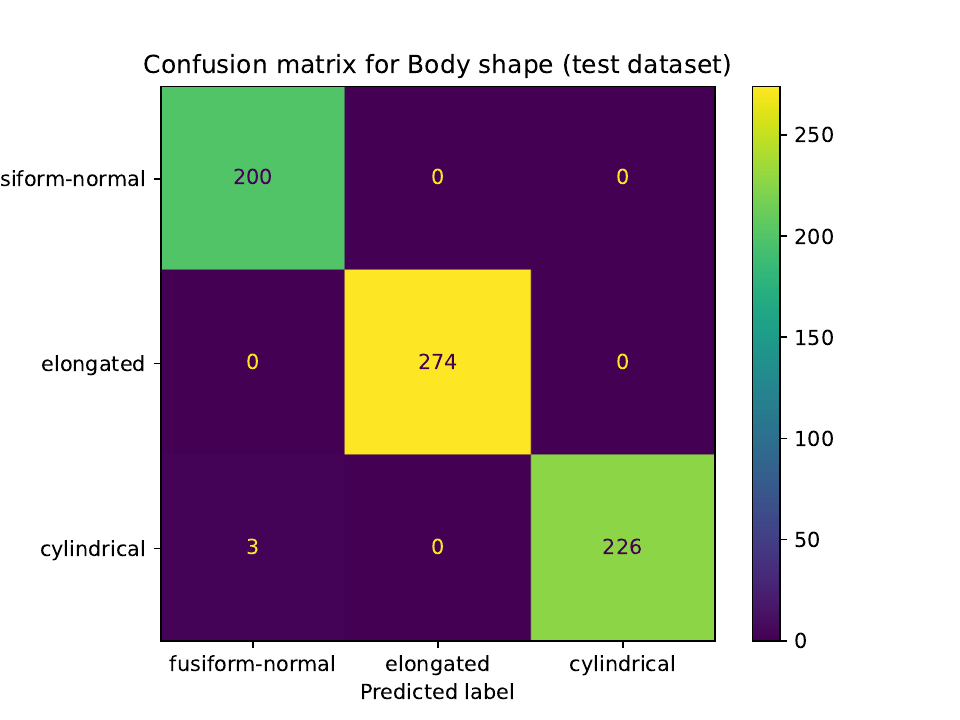}  \includegraphics[width=\linewidth]{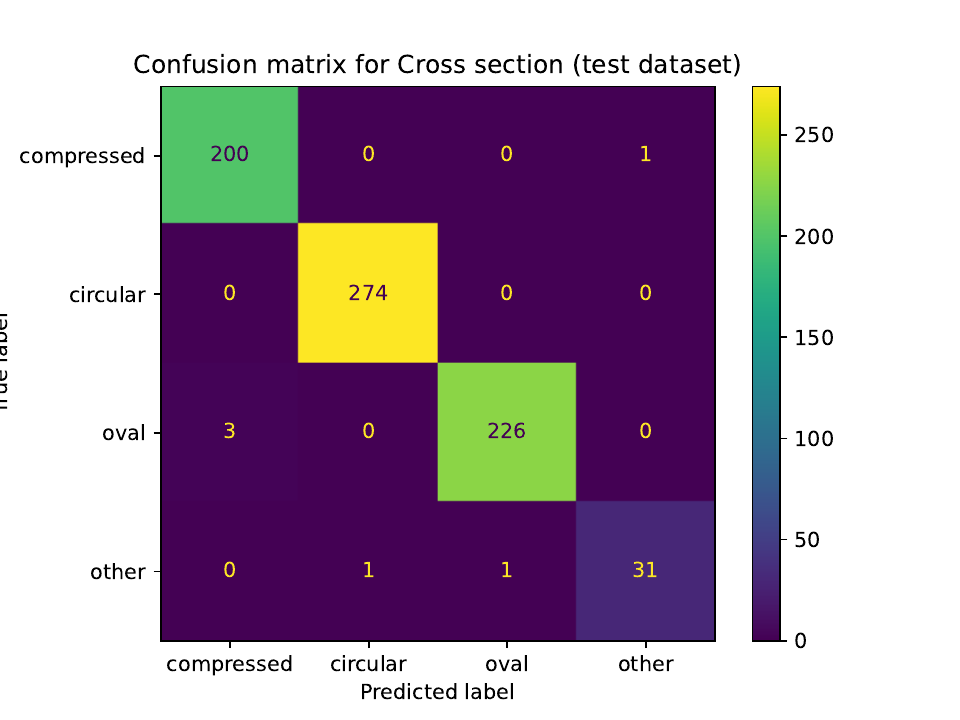}
  \caption{Classes prediction for 'Body shape' and 'Cross section'}
  \label{fig:classPredictedValues}
\end{figure}

\begin{figure}[t]
  \includegraphics[width=0.6\linewidth]{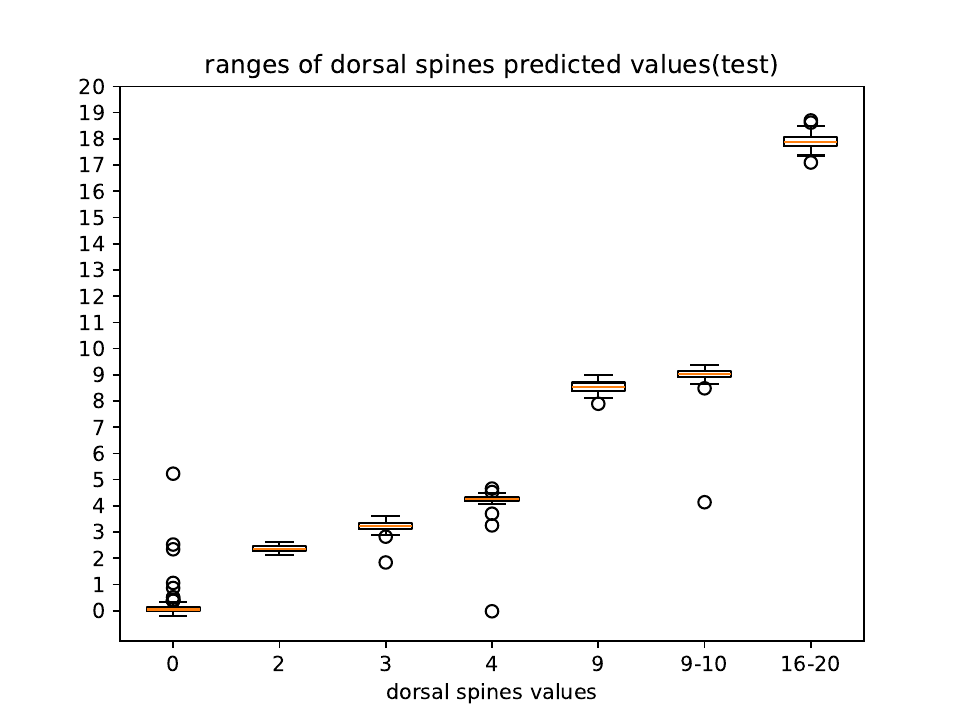}
  \includegraphics[width=0.6\linewidth]{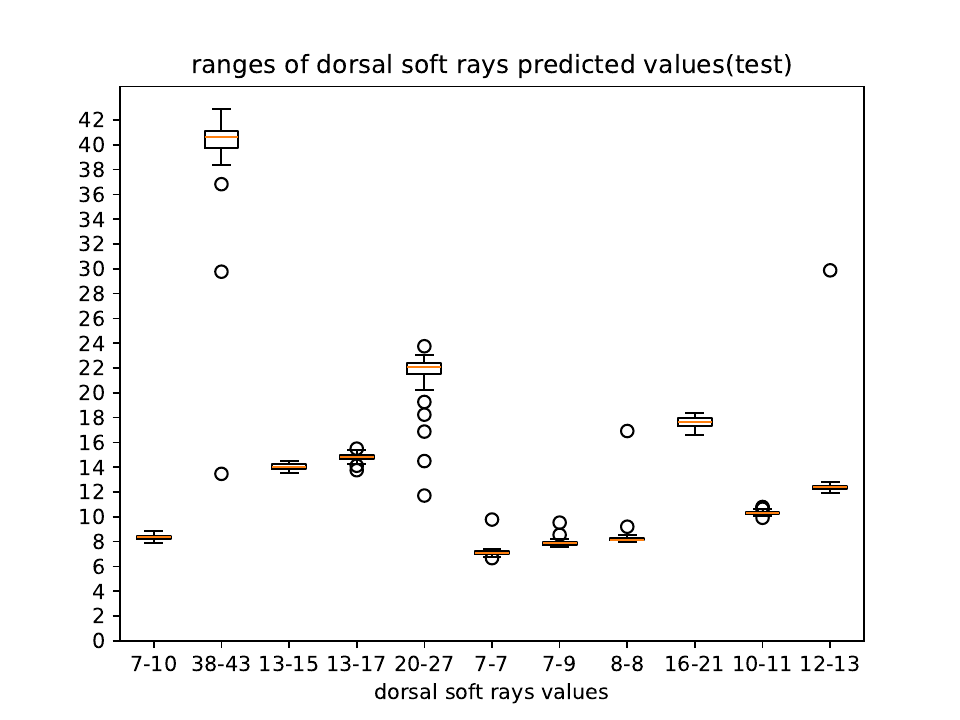}
  \includegraphics[width=0.6\linewidth]{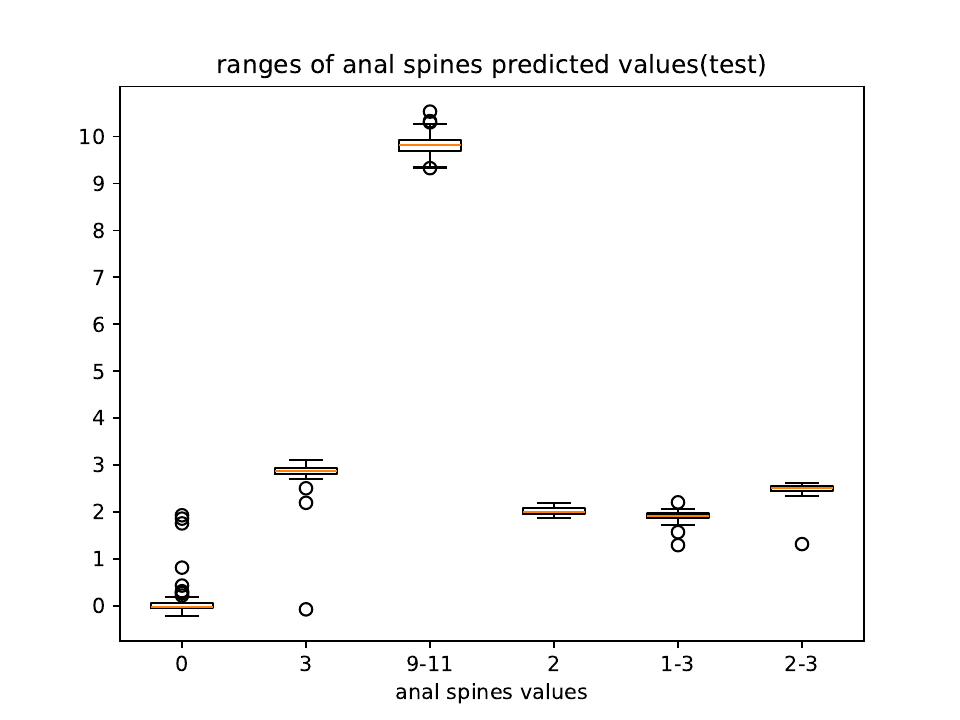}
  \includegraphics[width=0.6\linewidth]{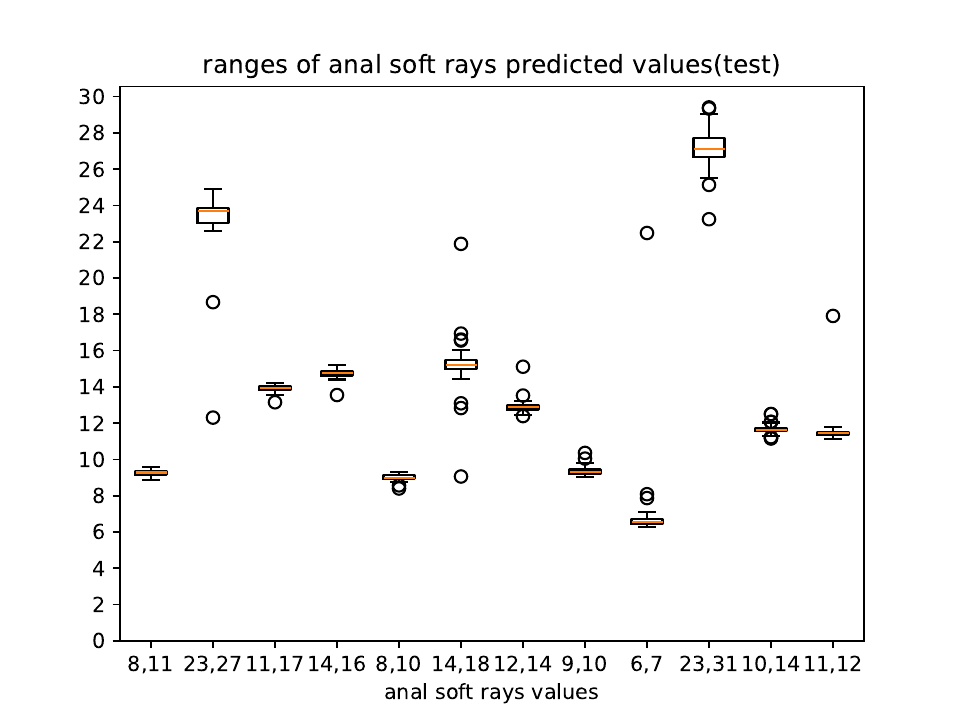}
  \caption{Ranges of predicted values for the test set. The box plot are using the true ranges of values even if some class share the same intervals.}
  \label{fig:rangePredictedValues}
\end{figure}

In figure~\ref{fig:rangePredictedValues} we verify that the predicted values of Dorsal/Anal Spines/Soft Rays fall into correct intervals on the test set. We also display the confusion matrices for the 2 categorical attributes: Body Shape and Cross Section on figure~\ref{fig:classPredictedValues}.

\section{Hardwood charcoal dataset}
\label{sec:presAIWOODdataset}

In this material, we add elements on the hardwood charcoal dataset. 
We introduce a hardwood charcoal dataset assembled with expert anthracologists within the AIWOOD ANR project~\footnote{https://anr.fr/Project-ANR-23-CE38-0013}, comprising $16$ species from $10$ genera and $3$ botanical families, $393$ individual trees and $3{,}313$ high-resolution micrographs in total with different original sizes (2880$\times$2048, 2559$\times$1917, 1023$\times$724, 1021$\times$745, 949$\times$744) (table~\ref{tab:dataset}).  Every specimen is documented in all three anatomical sections. Each species carries a complete expert IAWA annotation in the three-state regime.
This benchmark is substantially harder, and the difficulty is structural rather than merely a matter of cardinality: five of the ten genera contain more than one species (\emph{Prunus} $\times3$, \emph{Salix} $\times2$, \emph{Sorbus} $\times2$, \emph{Vaccinium} $\times2$), so a substantial fraction of the label space consists of congeneric pairs that share most of their gross anatomy. The dataset is also strongly imbalanced, from $16$ specimens (\emph{Arbutus unedo}, \emph{Rubus caesius}) to $39$ (\emph{Prunus avium}).

\begin{table}[!h]
\centering
\footnotesize
\setlength{\tabcolsep}{2.2pt}
\renewcommand{\arraystretch}{0.9}

\begin{tabular}{@{}p{1.0cm} p{3.0cm} c c c c c@{}}
\toprule
\multirow{2}{*}{Family} &
\multirow{2}{*}{Species} &
\multirow{2}{*}{\#Trees} &
\multicolumn{3}{c}{Images / section} &
\multirow{2}{*}{\#Img} \\
\cmidrule(lr){4-6}
& & & R & T & Tr & \\
\midrule

\multirow{6}{*}{Ericaceae}
& \textit{Arbutus unedo}            & 16 &  61 & 35 & 38 & 134 \\
& \textit{Calluna vulgaris}         & 31 & 130 & 66 & 72 & 268 \\
& \textit{Erica cinerea}            & 19 &  81 & 40 & 39 & 160 \\
& \textit{Rhodo. ferrugineum} & 20 &  67 & 45 & 44 & 156 \\
& \textit{Vaccinium myrtillus}      & 20 &  92 & 49 & 48 & 189 \\
& \textit{Vaccinium uliginosum}     & 19 &  88 & 42 & 42 & 172 \\

\midrule

\multirow{8}{*}{Rosaceae}
& \textit{Crataegus monogyna} & 36 & 156 & 79 & 82 & 317 \\
& \textit{Prunus avium}       & 39 & 111 & 64 & 78 & 253 \\
& \textit{Prunus persica}     & 26 & 161 & 84 & 85 & 330 \\
& \textit{Prunus spinosa}     & 35 & 149 & 65 & 76 & 290 \\
& \textit{Rosa canina}        & 21 &  92 & 42 & 45 & 179 \\
& \textit{Rubus caesius}      & 16 &  73 & 32 & 32 & 137 \\
& \textit{Sorbus aria}        & 22 &  70 & 44 & 53 & 167 \\
& \textit{Sorbus aucuparia}   & 29 & 124 & 63 & 66 & 253 \\

\midrule

\multirow{2}{*}{Salicaceae}
& \textit{Salix foetida}   & 24 & 72 & 48 & 48 & 168 \\
& \textit{Salix pentandra} & 20 & 60 & 40 & 40 & 140 \\

\midrule

\multicolumn{2}{@{}l}{\textbf{Total} (3 families, 10 genera, 16 species)}
& \textbf{393} & & & & \textbf{3,313} \\

\bottomrule
\end{tabular}

\caption{\textbf{Dataset composition.}
R: radial, T: tangential, Tr: transverse. All species are hardwoods and all carry a complete expert IAWA annotation (tables \ref{tab:intervals} and \ref{tab:annotations}).}
\label{tab:dataset}
\end{table}

\subsection{Expert charcoal hardwood criteria used in this work} 
Some of the anatomical criteria for fresh wood are available on Inside Wood database~\cite{wu2021wood}. However, since we are working with charcoal, the criteria are no longer the same due to the carbonization process; therefore, anthracology experts conducted a new study to provide us with criteria specific to charcoal. After cleaning, the following criteria were selected for the purposes of this work, available in table~\ref{tab:annotations}.

\begin{table}[ht]
\centering
\small
\setlength{\tabcolsep}{6pt}
% \begin{tabular}{@{}lccc@{}}
\begin{tabular}{@{}l|ccc|@{}}
\toprule
Species & $\mathcal{B}_1$ & $\mathcal{B}_2$ & $\mathcal{B}_3$ \\
\midrule
\emph{Arbutus unedo}            & 25, 26 & 40, 41     & 50     \\
\emph{Calluna vulgaris}         & 24     & 40, 41     & 50     \\
\emph{Crataegus monogyna}       & 24, 25 & 40         & 50     \\
\emph{Erica cinerea}            & 25     & 40, 41     & 49, 50 \\
\emph{Prunus avium}             & 25, 26 & 41         & 49, 50 \\
\emph{Prunus persica}           & 25, 26 & 40, 41     & 49, 50 \\
\emph{Prunus spinosa}           & 25, 26 & 40, 41     & 50     \\
\emph{Rhododendron ferrugineum} & 25     & 40         & 50     \\
\emph{Rosa canina}              & 25, 26 & 40, 41, 42 & 49, 50 \\
\emph{Rubus caesius}            & 25     & 41, 42, 43 & 48, 49 \\
\emph{Salix foetida}            & 26, 27 & 40, 41     & 50     \\
\emph{Salix pentandra}          & 26, 27 & 40, 41     & 50     \\
\emph{Sorbus aria}              & 24, 25 & 40         & 50     \\
\emph{Sorbus aucuparia}         & 24, 25 & 40         & 50     \\
\emph{Vaccinium myrtillus}      & 25, 26 & 40, 41     & 50     \\
\emph{Vaccinium uliginosum}     & 25, 26 & 40, 41     & 50     \\
\bottomrule
\end{tabular}
\caption{\textbf{Expert-annotated admissible intervals for the three ordinal
blocks.} $\mathcal{B}_1$: criteria 24--27 ; $\mathcal{B}_2$: criteria 40--43 ; $\mathcal{B}_3$: criteria 46--50. Full text definitions of each criteria and blocks are given in the IAWA hardwood list~\cite{xian2018zero} and InsideWood~~\cite{wu2021wood}. Criteria declared variable are
included, so each entry is the widest band consistent with the annotation. The
bands overlap heavily, and several species are identical across all three blocks which is why the ordinal heads collapse onto a constant.}
\label{tab:intervals}
\end{table}

The criteria for the three intervals blocks are available in table~\ref{tab:intervals}.

\begin{table*}[t]
\centering\small
\begin{tabular}{@{}l p{0.80\textwidth}@{}}
\toprule
Species & Charcoal IAWA criteria \\
\midrule
\emph{Arbutus unedo} & 1 4v 5 7v 9 10v 12v 13 14v 21 25 26 30 36 37 40 41 50 53 60 61 62 63 64 65 66 69 72 75 76v 78 92 93 98 103v106 107 115 136v 140v 142v 143v \\[2pt]
\emph{Calluna vulgaris} & 1 5 9 13 14v 21 24 30 31v 40 41 50 52 58 62 63 66 69 71 76 91v 92 93 96 105 116 \\[2pt]
\emph{Crataegus monogyna} & 1 4v 5 9 13 20v 24 25v 30 36 37v 38 40 50 53 58v 62 66 69 70 71 76 77 78 86v 93 97 104 106v 115 136v 144v \\[2pt]
\emph{Erica cinerea} & 1 5 9 12 13 14v 21 25 30 40 41 49 50 53 56v 62 66 69 71 76 78 91 92 97 98 103 106 107 114 115 \\[2pt]
\emph{Prunus avium} & 1 4 5 10v 11 12v 13 25 26 30 36 37 41 49 50v 53 58 61v 62 66 69 70 76 97 98v 103 105 106 107v 115 136v 144v \\[2pt]
\emph{Prunus persica} & 1 3 4 10 11 12v 13 25 26 30 36 37 40 41 49v 50 53 58 61v 62 66 69v 70 71 72 76 92 93 97v 98 103 105 106 107v 115 136v 144v \\[2pt]
\emph{Prunus spinosa} & 1 4 5v 10v 11 12v 13 22 25 26 30 36 37 40 41 50 53 58 61v 62 66 69v 76 92 93 97v 98 103 105 106 107v 115 136v 144v \\[2pt]
\emph{Rhododendron ferrugineum} & 1 2v 5 9 12 14 20 21 25 30 40 50 52 53 62 63 66 69 70 76 90 91 92 97 98 103 107 108 116 \\[2pt]
\emph{Rosa canina} & 1 3 4 9 12v 13 21 23v 25 26 30 36 37 40 41 42v 49 50 52 53 62 63 64 66 70 71 72 76 78 91 92 93 97 98 99 102 103 105 107v 108 109v 110 115 116 136 \\[2pt]
\emph{Rubus caesius} & 1v 2v 4v 5 9 11 13 25 30 41 42 43v 45 48 49v 52 53 62 63v 66 75 76 91v 92 97 98 102 103 105 115 120v \\[2pt]
\emph{Salix foetida} & 1 4 5 6 9 10 13 20 23 26 27 40 41v 50 61 75v 76v 89 91 92 96 106 107 115v 116 \\[2pt]
\emph{Salix pentandra} & 1 4v 5 6v 7v 9 10 13 20 23 26 27 40 41v 50 56 61 75v 76v 89 91 92 96 106 107 115v 116 \\[2pt]
\emph{Sorbus aria} & 1 4v 5 9 13 20v 24 25v 30 36 37v 38 40 50 53 58v 62 66 69 70 71 76 77 78 86v 93 97 104 106v 115 136v 138 144v 145v 146v \\[2pt]
\emph{Sorbus aucuparia} & 1 4v 5 9 12v 13 20v 24 25v 30 36 37v 38 40 50 53 58v 62 65 69 70 71 76 77 78 86v 93 97 104 106v 115 136v 138v 144v 145v 146v \\[2pt]
\emph{Vaccinium myrtillus} & 1 2v 5 6 9 12 13v 14 15 20 21v 25 26 31 40 41 50 53 62 63 66 70 75v 76 97 98v 102 103 105 107v 108v 110v 116 120v \\[2pt]
\emph{Vaccinium uliginosum} & 1 2v 5 6 9 12 13v 14 15 20 21v 25 26 31 40 41 50 53 62 63 66 70 75v 76 97 98v 102 103 105 107v 108v 110v 116 120v \\[2pt]
\bottomrule
\end{tabular}
\caption{\textbf{Expert IAWA annotations for the 16 species, (73 binary and 13 used for the three intervals blocks table~\ref{tab:intervals}).} Each entry lists the microscopic feature numbers of the IAWA hardwood list~\cite{xian2018zero} recorded for that species on carbonised material. Numbers in \textit{v} are declared \emph{variable} within the species. Criteria absent from an entry are absent in that species.}
\label{tab:annotations}
\end{table*}

\subsection{Implementation details for the hardwood charcoal dataset}
\label{sec:impl}

Micrographs are converted to grayscale and tiled into non-overlapping $500\!\times\!500$ crops, resized to the native input resolution of the backbone and then normalised with its ImageNet statistics. The backbone is frozen throughout.

The set encoder uses $d_h=d=D$, max pooling, weights shared across sections, and dropout $p=0.6$. Training uses AdamW~~\cite{martins2012database} with learning rate $10^{-3}$, weight decay $10^{-4}$ and batch size $32$, for at most $200$ epochs, with \texttt{ReduceLROnPlateau} on the validation loss (factor $0.5$, patience $3$, minimum $10^{-5}$) and early stopping with patience $10$; the parameters of the best validation epoch are restored. The uncertainty parameters $\bm{s}$ are optimised in the same parameter group as the network.

\subsection{Further analyses on the predicted IAWA criteria}

\begin{figure}[t]
\centering
\includegraphics[width=\columnwidth]{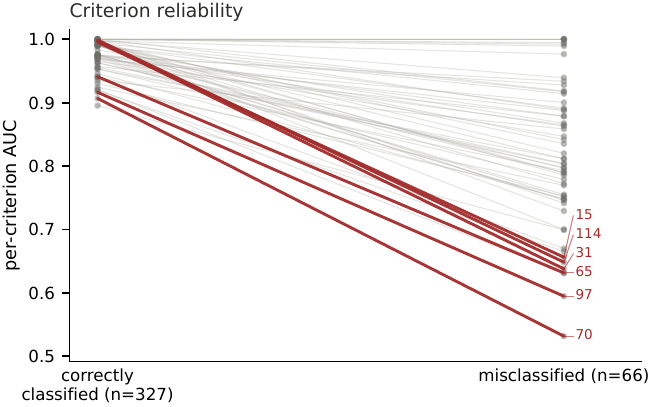}
\caption{Per-criterion AUC on correctly classified specimens against misclassified ones. Every scorable criterion declines; the six largest drops are highlighted and labelled by IAWA number.}
\label{fig:inter-auc}
\end{figure}

Figure~\ref{fig:inter-auc} shows how criterion reliability changes between correctly and incorrectly classified specimens. Two features are worth noting. First, the decline is universal: not one of the scorable criteria is predicted better on errors than on correct identifications, so the degradation is systematic rather than the effect of a few outliers. Second, it is concentrated: most criteria remain above $0.80$ on misclassified specimens, and six fall below $0.75$ (on right side). A failed identification is therefore not a wholesale misreading of the anatomy but the misreading of a small, identifiable subset of features, which is what makes the predicted description usable as an audit trail. That these six concern wall thickness, cell dimensions and fine pitting is consistent with the effect of carbonisation on quantitative anatomical characters~\cite{zemke2025variation}.

\end{document}